\documentclass[conference]{IEEEtran}
\IEEEoverridecommandlockouts
\usepackage{multicol}
\usepackage{multirow}
\usepackage{url}

\usepackage{cite}
\usepackage{amsmath,amssymb,amsfonts}
\usepackage{algorithmic}
\usepackage{graphicx}
\usepackage{textcomp}
\usepackage{xcolor}
\def\BibTeX{{\rm B\kern-.05em{\sc i\kern-.025em b}\kern-.08em
    T\kern-.1667em\lower.7ex\hbox{E}\kern-.125emX}}
\begin{document}

\title{Chain-of-Interaction: Enhancing Large Language Models for Psychiatric Behavior Understanding by Dyadic Contexts}

\author{\IEEEauthorblockN{Guangzeng Han\IEEEauthorrefmark{1}, Weisi Liu\IEEEauthorrefmark{1}, Xiaolei Huang\IEEEauthorrefmark{1}, Brian Borsari\IEEEauthorrefmark{2}\IEEEauthorrefmark{3}}
\IEEEauthorblockA{\IEEEauthorrefmark{1}Department of Computer Science,
University of Memphis, Memphis, United States\\
\texttt{\{ghan,wliu9,xiaolei.huang\}@memphis.edu}} 
\IEEEauthorblockA{\IEEEauthorrefmark{2}San Francisco Veterans Affairs Health Care System, San Francisco, United States\\
\texttt{brian.borsari@va.gov}	}
\IEEEauthorblockA{\IEEEauthorrefmark{3}Department of Psychiatry and Behavioral Sciences, University of California, San Francisco, United States\\
}}

\maketitle

\begin{abstract}

Automatic coding patient behaviors is essential to support decision making for psychotherapists during the motivational interviewing (MI), a collaborative communication intervention approach to address psychiatric issues, such as alcohol and drug addiction.
While the behavior coding task has rapidly adapted language models to predict patient states during the MI sessions, lacking of domain-specific knowledge and overlooking patient-therapist interactions are major challenges in developing and deploying those models in real practice.
To encounter those challenges, we introduce the Chain-of-Interaction (CoI) prompting method aiming to contextualize large language models (LLMs) for psychiatric decision support by the dyadic interactions.
The CoI prompting approach systematically breaks down the coding task into three key reasoning steps, extract patient engagement, learn therapist question strategies, and integrates dyadic interactions between patients and therapists. 
This approach enables large language models to leverage the coding scheme, patient state, and domain knowledge for patient behavioral coding.
Experiments on real-world datasets can prove the effectiveness and flexibility of our prompting method with multiple state-of-the-art LLMs over existing prompting baselines. 
We have conducted extensive ablation analysis and demonstrate the critical role of dyadic interactions in applying LLMs for psychotherapy behavior understanding.\footnote{Code available at~\url{https://github.com/trust-nlp/CoI-Psychotherapy}}

\end{abstract}
\IEEEpeerreviewmaketitle

\section{Introduction}



\begin{figure}[tbh]
\centering
\includegraphics[width=0.48\textwidth]{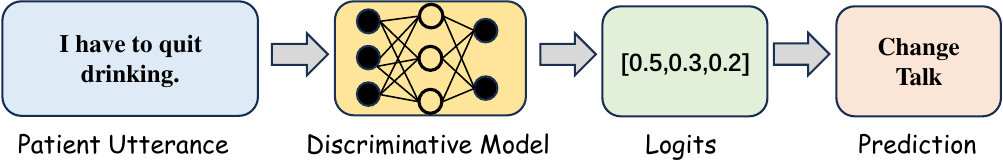}  
\caption{Diagram illustrating automatic behavioral coding methods based on discriminative models.}
\label{fig1:NN-based method}
\end{figure}

Motivational interviewing (MI)~\cite{miller2023motivational} is defined as ``particular way of talking with people about change and growth to strengthen their own motivation and commitment.'' MI can facilitate change in behaviors associated with mental health issues, which are among the most complex health problems impacting over one billion people worldwide~\cite{world2022world}. The Motivational Interviewing Skill Code (MISC)~\cite{miller2003miti} is a coding system that provides comprehensive and structured coding schema for examining therapist and patient behaviors (utterances) and assessing different aspects of the intervention (e.g., therapist empathy).
However, MI session assessments require significant time commitment, labor costs, and professional training~\cite{atkins2014scaling}, which may not meet immediate training or clinical needs.
An alternative solution is to build machine learning models to automate coding process in Fig.~\ref{fig1:NN-based method}.


Fig.~\ref{fig1:NN-based method} presents current standard process of automatically coding MI sessions that first vectorize patient utterances into neural representations and then feed the representations to discriminative classifiers for final predictions~\cite{TANANA2016Comparison}.
For example, recent studies have developed MI classifiers by recurrent neural networks (e.g., LSTM or GRU)~\cite{ huang2018modeling, cao2019observing} and BERT~\cite{devlin2019bert, Tavabi2020Multimodal} for automatically predicting MISC codes at utterance and session levels, which have achieved promising performance. 
However, a major issue of the existing ML classifiers is missing incorporating the psychiatric domain knowledge, including MISC coding manuals, which guide and train human professionals to annotate patient behaviors.
The existing methods require costly and time-consuming human annotations and may not integrate domain-specific knowledge from the MISC coding schema by only feeding utterances into classifiers for predictions, such as MISC concepts and definitions used by human experts in behavioral coding. 
Thus, a concrete question is: \textit{how can we instruct models to annotate MI sessions following the MISC schema like psychiatric professionals}?

\begin{table*}[htp]
\caption{Samples of MISC-coded data. The same sub-codes may be encoded as different patient codes due to their different valence (+/-)}
\resizebox{\textwidth}{!}{%

\begin{tabular}{c|l|l|l|l}
\multirow{2}{*}{Patient code}   & \multirow{2}{*}{Sub-code} & \multirow{2}{*}{Description}            & \multirow{2}{*}{Therapist utterance (Code)}         & \multirow{2}{*}{Patient utterance} \\
                                &                           &                                         &                                                     &                                    \\ \hline \hline
\multirow{4}{*}{Change Talk}    & Commitment+               & Patient makes a changing commitment.    & Do you want to change? (Closed Question)            & I am going to stop smoking.        \\ \cline{2-5} 
                                & Ability+                  & Assess client’s ability to change.       & You’ve been through a lot. (Support)                & I can do it.                       \\ \cline{2-5} 
                                & Desire+                   & A desire to alter the target behavior.   & I want to talk about your motivation.(Structure )   & I want to quit doing drugs.        \\ \cline{2-5} 
                                & Reason+                   & Reasons for changing the behavior       & What are some reasons for quitting? (Open Question) & I’m killing myself.                \\ \hline
\multirow{3}{*}{Follow/Neutral} & N/A                       & Reporting information.                   & How often do you drink? (Closed Question)           & Usually 4-5 days.                  \\ \cline{2-5} 
                                & N/A                       & Follow therapist.                       & Do you have children? (Closed Question)             & I have three daughters.            \\ \cline{2-5} 
                                & N/A                       & Ask question.                            & The support group meets from 4 to 5 pm. (Give Information)  & Do you know when they meets?       \\ \hline
\multirow{4}{*}{Sustain Talk}   & Commitment-               & Patient makes a maintaining commitment. & Speeding will cost you your license. (Warn)         & I will never slow down when I drive.                 \\ \cline{2-5} 
                                & Ability-                  & Assess client’s ability to change.      & Only you know what’s best for you. (Emphasize Control)   & I cannot stop overeating even when I try.     \\ \cline{2-5} 
                                & Desire-                   & A desire to maintain the  behavior.      & If you get bored you’ll use drugs. (Warn)           & I want to keep getting high.       \\ \cline{2-5} 
                                & Reason-                   & Reasons for maintaining the behavior.    & Put your health first! (Direct)                      & At least I get relaxed when I drink. \\ \hline
\end{tabular}}
\label{table,MISC-codes-sample}
\centering
\end{table*}

Large language models (LLMs), such as GPT~\cite{openai2023gpt4, ouyang2022instructGPT} and Llama2~\cite{touvron2023llama}, has undergone a rapid evolution in recent years and have demonstrated a significant potential transform mental health and psychotherapy interventions, such as depression detection in social media~\cite{yang2023mentalllama, ji2022mentalbert} and psychotherapy chatbot~\cite{liu2023Chatcounselor}.
Effective prompting design for LLMs is the key to achieve precise diagnosis and understanding of patient behaviors. 
Prompting methods involve how we put together and format questions or commands for the model to instruct generating specific responses. 
For example, prompting methods like Chain of Thought (CoT)~\cite{wei2022chain}, Thread of Thought~\cite{zhou2023thread}, and Retrieval-Augmented Generation (RAG)~\cite{zhang2023optimizing,jing2024large} can effectively solve common sense reasoning tasks –– by splitting complex questions into several easier and key parts of general questions. 
However, these kinds of prompting techniques do not consider the dyadic contexts of MI sessions, interactions between patients and therapists.
The dyadic interaction between psychotherapists and patients is a natural characteristic of MI sessions, which is commonly overlooked in the existing prompt approaches.
As the interactive context can provide key insights for accurately inferring patient behaviors, the challenges call for new methods to integrate psychiatric knowledge, understand dyadic contexts, and equip with strong reasoning capability.

To address these specific challenges, we propose the \textit{Chain-of-Interaction} prompting method aiming to incorporate domain knowledge from MISC coding schema and in-session interactions. 
It utilizes a series of prompt stages to enhance the reasoning abilities of LLMs in the task of coding patient utterance, drawing upon psychological domain knowledge and the style of interactions between patients and therapists from three consecutive stages: Interaction Definition, Involvement Assessment, and Valence Analysis.
The Interaction Definition stage aims to help LLMs define the therapist-patient interactions by categorizing the unique MISC codes of each utterance.
Next, the Involvement Assessment evaluates patient engagement by assessing willingness for self-exploration and emotional expression.
Finally, Valence Analysis stage is to integrate the general sentiment of patient and the clues from previous two stages, we expect LLMs can utilize the integrated information to mimic the reasoning process of human experts.

We conduct extensive experiments on datasets derived from real-world MI sessions addressing alcohol usage disorder.
We examine three prompting baselines and experiment with four state-of-the-art auto-regressive LLMs, including Llama2~\cite{touvron2023llama}, Falcon~\cite{almazrouei2023falcon}, Mistral~\cite{jiang2023mistral}, and ChatGPT~\cite{openai2022chatgpt}.
Our experiments demonstrate the effectiveness of the Chain-of-Interaction (CoI) prompting method, which can outperform multiple baselines by a large margin. 
The consistent improvements in Macro-F1 and Micro-F1 scores can indicate that CoI may enhance LLMs under data and label imbalance.
Furthermore, our ablation study reveals the critical contribution of domain knowledge and dyadic interactions in LLM performance.

\begin{figure*}[htp]
\centering
\includegraphics[width=1\textwidth]{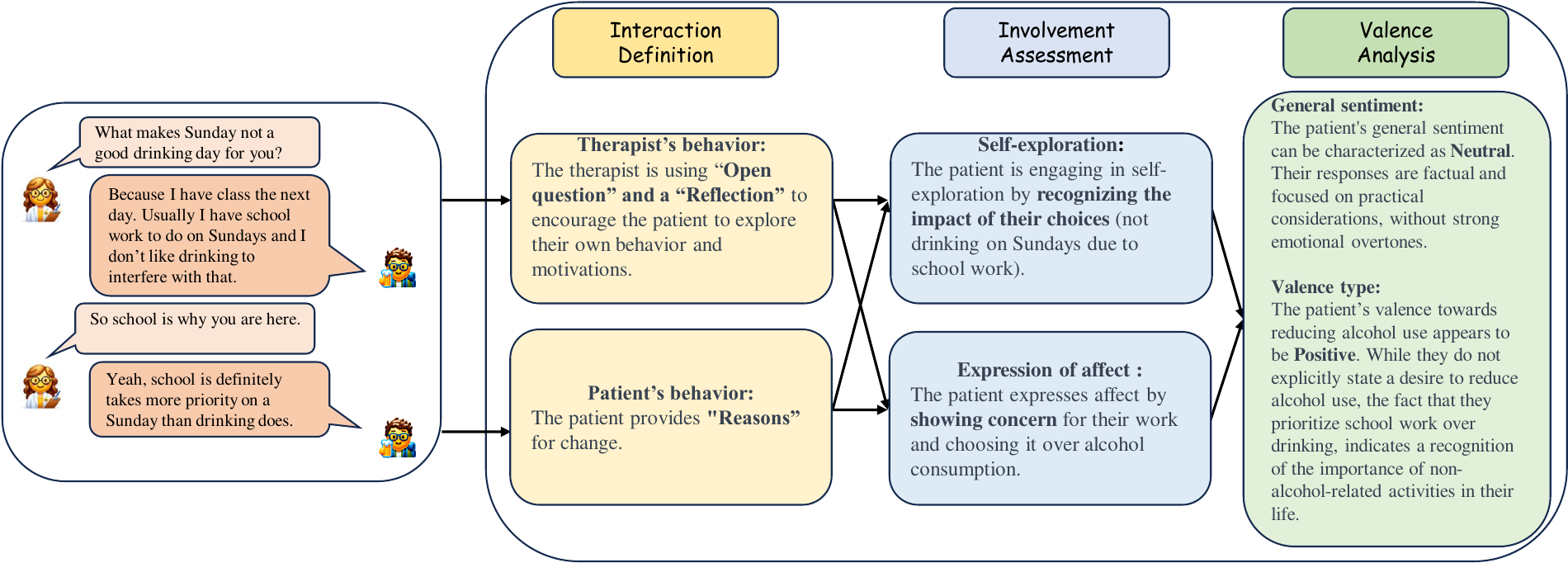}  
\caption{Overview of the ``Chain-of-Interaction'' prompting method, we use three colors to denote different stages.}
\label{figure2:CoI-diagram}
\end{figure*}

\section{Data}

We used sessions from two independent randomized clinical trials~\cite{borsari2015session}, which conducted MI sessions with 249 university students who had been mandated for alcohol-related infractions (e.g., possession, vandalism). 
Each MI session lasts between 45 to 60 minutes and were audio recorded.
To protect user privacy and identities, these recorded sessions were transcribed into text, de-identified, and coded using the Motivational Interviewing Skills Code (MISC version 2.0)~\cite{miller2008manual}. 
MISC encodes patient utterances into three mutually exclusive overarching categories: Follow/Neutral, Change Talk, and Sustain Talk. 
``Follow/Neutral'' encompasses responses where the patient aligns with, remains neutral, or seeks information from the therapy without directly addressing behavior change; ``Change Talk'' involves statements indicating an intention to change the behavior; and ``Sustain Talk'' includes utterances that suggest maintaining the current behavior.
Change language can be conceptualized as being on a continuum with a positive valence indicating Change Talk, a negative valence representing Sustain Talk, and a neutral valence corresponding to Follow/Neutral.


To control data quality, this study did not consider therapy records that had annotation errors or incomplete annotations.
Per recording session, we lower-cased utterances and extracted data segments consisting of every ten non-overlapping utterances as one data entry, aiming to provide dyadic interactions between patients and therapists. 
For each data entry, we utilized the MISC code of the last patient utterance of the entry as the ground truth label. 
This method ensures that the model's focus is on the patient's most recent utterance, accurately reflecting their behavioral state at that particular moment in the therapy session.
We summarize data statistics in Table~\ref{table: Data_statistics} for raw and processed data to allow for replication.

\begin{table}[ht]
\caption{Data statistics. $\overline{\mbox{L}}$ refers to the average number of tokens per utterance. $\overline{\mbox{U}}$ is the average utterance count per therapy session.
}
\centering
\begin{tabular}{c|c|c|c|c}
Utterance & Change Talk & Follow/Neutral & Sustain Talk & $\overline{\mbox{L}}$ \\ \hline \hline
48,529 & 13,298 & 29,025 & 6,026 & 30.2  \\ 
Session & Female & Male & Therapist & $\overline{\mbox{U}}$\\ \hline\hline
249 & 95 & 154 &14  & 194.9
\end{tabular}
\label{table: Data_statistics}
\end{table}

In this study, we utilize the three MISC client language categories for the auto behavior coding task, and we present examples of the MISC codes and utterances in Table~\ref{table,MISC-codes-sample}.
The examples show interactive and contextual effects exist between patients and therapists that patient utterances can shape therapeutic strategies for interventionists and in turn the conversation engagement will impact on patient behaviors.
While past studies\cite{cao2019observing, Tavabi2020Multimodal} have shown concatenating patient utterance and its previous utterances can improve prediction accuracy of the MISC coding task, how the patient-therapist interactions can inform model predictions is underexplored.
To effectively utilize this information, we proposed the Chain-of-Interaction (CoI) prompting, which helps large language models in automatically coding motivational interview patient behavior by leveraging the interactions between patients and therapists, as well as the engagement of patients during counseling sessions.

\section{The Chain-of-Interaction Prompting}
\begin{figure*}[htp]
\centering
\includegraphics[width=0.97\textwidth]{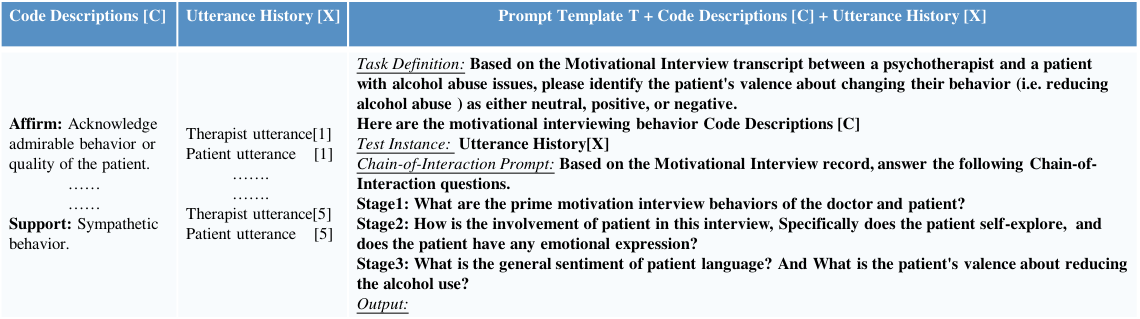}  
\caption{Details of the prompt design for the ``Chain-of-Interaction'' method.}
\label{figure3:-Prompt-detail}
\end{figure*}


MI efficacy theory~\cite{miller2009toward,Magill2019} suggests that dyadic interactions play a critical role in therapist skills, patient behaviors, and outcomes –– guide and inspire the development of our new prompting method.
Two aspects of this dyadic interaction are linked to MI efficacy. 
First, there is a technical component in which therapist MI Consistent (MICO) utterances can selectively elicit and strengthen client change language which is predictive of subsequent behavior change. 
There is also a relational component which incorporates the interpersonal aspects of the therapist and client relationship such as MI Spirit and empathy.
We propose the \textit{Chain-of-Interaction} prompting, aiming to incorporate the psychological domain knowledge and the two aspects of therapist-patient interaction to promote the contextual awareness of large language models in the behavioral coding task.
Fig.~\ref{figure2:CoI-diagram} illustrates our prompting framework of Chain-of-Interaction with three continuous stages: 1) Interaction Definition, 2) Involvement Assessment, 3) Valence Analysis.
Through this structured prompting approach, we imbue LLMs with a new level of understanding and analytical depth that parallels professional psychological expertise.

First, the Interaction Definition stage enables the LLMs learning MI domain knowledge and objectively understanding the interaction patterns.
Next, the Involvement Assessment stage inquiries LLMs to examine emotional and self-expression clues from the patient utterances to infer the patient's engagement level, aiming to bridge correlations between patient mental states and their behavior outcomes. 
Finally, in the Valence Analysis stage, LLMs predict the patient's utterance valence by integrating information about the patient's emotional state and interaction pattern.
Through this structured approach, we imbue LLMs with a level of understanding and analytical depth that parallels professional psychological expertise.
More details about the Chain-of-Interaction prompting methods can be found in Fig.~\ref{figure3:-Prompt-detail}.

\subsection{Interaction Definition}

Our goal is to equip predictive models with strong reasoning capabilities as human professionals, while concatenating dyadic interactions and patient utterances as features is the major approach for existing end-to-end MISC classifiers~\cite{mishra2023e, tran2023multimodal, Tavabi2020Multimodal, cao2019observing, huang2018modeling}. 
Unlike the feature integration approach, human professionals ideally utilize MICO strategies to interact with patients and the MISC coding schema can be used to annotate patient behaviors, which is a fundamental difference.
To encounter the wide-existing yet unsolved issue, we introduce the \textit{Interaction Definition} stage, which allows Large Language Models (LLMs) to understand and interpret the interactive contexts of therapists and patients by the guidance of MISC behavior coding schema and definitions.

We design instructions to guide LLMs understanding interactions between therapists and patients and mental states of patients via two major parts of prompt modules, 1) definitions of MISC codes and coping with 2) dyadic contexts.
The instructional prompts aim to inspire LLMs to capture MI domain knowledge and master the MISC coding schema.
To achieve this, we start with feeding LLMs with extracted MI definitions from MISC coding manuals~\cite{miller2008manual, miller2023motivational}.
After the initial instruction, we then ask LLMs to review a patient utterance with its dyadic contexts. 
The process is to enable LLMs behave like human professionals that apply domain knowledge by their in-context learning ability.
We request LLMs to attempt annotating utterances of the dyadic context by the MISC therapist MICO categories (e.g. open-ended questions, affirmation, reflections) and the patient utterance subcategories shown in Table~\ref{table,MISC-codes-sample}. 
We illustrate examples of the process and prompt examples in Fig.~\ref{figure2:CoI-diagram} and \ref{figure3:-Prompt-detail}.
For example, Fig.~\ref{figure2:CoI-diagram} shows that LLMs encode the patient and therapist utterances as ``Reason'' and ``Open question'', respectively, which will help LLMs understand patient dynamic mental states and provide more information for the following stages.
The stage meets our initial goal by instructing LLMs by the behavior coding task, incorporating domain knowledge as instructions, and equipping LLMs with domain-specific reasoning capabilities.



\subsection{Involvement Assessment}

MI coding teams objectively code MISC patient behavior while also subjectively rating the relational aspects of the session\cite{borsari2015session} (e.g., engagement).
Prior research~\cite{gagneur2020respiratory} has also demonstrated that patients who are highly engaged in treatment have more Change Talk, and less Sustain Talk, therefore assessing patient engagement will help the model better understand patient tendencies regrading target behavior.
To utilize these auxiliary subjective paths to help the objective behavioral coding, we propose the \textit{Involvement Assessment} stage, which makes LLMs to mimic the rating process of the MI coders by assessing the involvement of patient.

In this stage, we formulate prompts to remind LLMs to revisit the original dyadic contexts and the Interaction Definitions (i.e. outputs of \textit{Interaction Definition stage}), extracting two important cues about engagement from them: self-exploration and emotional expression~\cite{rollnick2008motivational,steven2023using}.
These prompts aims to guide LLMs to emulate the rating process professionals use when coding MI patient behavior. 
To aid in simulating this process, we first require LLMs to consider not only the original dyadic contexts but also the Interaction Definitions of stage 1 outputs. 
We then prompt LLMs to assess patient engagement through aspects of self-exploration and emotional expression.
The example in Fig.~\ref{figure2:CoI-diagram} illustrates how the LLM interprets specific methods of self-exploration and emotional expression based on the ``reason'' behavior of patients from the first stage.

From the clear interaction patterns obtained in the \textit{Interaction Definition} stage to the subjective assessment of engagement in this stage, LLMs can benefit from domain-specific knowledge and emulate the thinking habits of professionals.
Furthermore, the emotional expressions identified in this stage serve as a foundation for the analysis in the final stage.


\subsection{Valence Analysis}


In MI, valence reflects the client’s utterances regarding changing the target behavior.
A positive valence indicates Change Talk, a negative valence represents Sustain Talk, and a neutral valence corresponds to Follow/Neutral.
According to their correspondence, we transform the behavioral coding task into valence coding task.
Valence is a component of sentiment, focusing specifically on the positive-negative axis of emotional response. 
Sentiment, however, is a more comprehensive term that captures the full spectrum of emotional states and attitudes, including valence as one of its dimensions.
Based on this characterization, we propose the \textit{Valence Analysis} stage, which aims to help LLMs better perform the valence coding task through two sub-stages.
The first sub-stage involves utilizing general sentiment analysis to aid LLMs in comprehending sentimental states of patients. 
The second one integrates outputs obtained from all three stages to make LLMs to more comprehensively utilize the domain-specific knowledge and the interactive contexts.

In this stage, we start with instructing LLMs to perform a generic sentiment analysis based on the original utterances and the emotional expressions obtained from the \textit{Involvement Assessment} stage.
The goal is to allow LLMs to learn more about the patient's sentimental state before reasoning about our final task objective: Valence, through a simpler and more generalized task.
After the sentiment analysis, we finally prompt LLMs to review the outputs obtained through the three progressive stages, which include the required objective criteria in the MISC coding schema, as well as the subjective rating that professionals would refer to when coding, and the sentimental state of the patient's language.
By integrating the output from these three stages, LLMs not only understood the pattern of interaction between the patient and therapist, but also assessed the patient's level of engagement as well as analyzed the patient's emotional state.
These procedures help LLMs to reason and code more rationally about the patient's Valence by utilizing domain knowledge and patient-therapist interactions as professional coders do.


\section{Experiment}

To examine the effectiveness of our proposed `Chain-of-Interaction' (CoI) prompting method, we conducted a comparative analysis with state-of-the-art baselines and performed a detailed ablation study.
We set our experimental results against three established baseline methods: Zero-Shot prompting~\cite{wang2019survey,yin2019benchmarking}, Few-Shot prompting ~\cite{brown2020language,liu2023large}, and Zero-Shot Chain of Thought (ZeroCoT)~\cite{kojima2022large} prompting.
In addition, our work includes an ablation study focusing on the proposed Chain-of-Interaction (CoI) method, where we removed each individual stage within the CoI. 
By assessing the performance resulting from the removal of each stage, we measured its contribution to the overall effectiveness.
We selected four representative auto-regressive LLMs to perform the experiments, including Llama2-13B-Chat~\cite{touvron2023llama}, Falcon-7B-Instruct~\cite{almazrouei2023falcon}, Mistral-7B-Instruct~\cite{jiang2023mistral}, and ChatGPT~\cite{openai2022chatgpt}.
We used model-specific official prompt templates and followed each model's default sampling strategies. 
Our assessment employed Micro-F1 score to gauge overall model accuracy and Macro-F1 score to ensure fairness across different classes, offering a balanced evaluation of the models' performance in varied environments.

\subsection{Baselines}
To demonstrate the effectiveness of our proposed Chain-of-Interaction prompting method, we compare it with three state-of-the-art prompting techniques: 1) Zero-Shot Prompting~\cite{wang2019survey,yin2019benchmarking}, 2) Few-Shot Prompting~\cite{brown2020language,liu2023large}, 3) Zero-Shot Chain-of-Thought (ZeroCoT) Prompting~\cite{kojima2022large}.

\subsubsection{Zero-Shot Prompting}

Zero-shot prompting~\cite{wang2019survey,yin2019benchmarking}, where LLMs receive only a sample and task description without corresponding examples or specially designed prompts, may be insufficient for complex and domain-specific tasks.
This approach relies entirely on the LLMs' inherent knowledge and adaptability, potentially causing it to bypass important reasoning steps and overlook critical information when lacking necessary context.
In MI behavioral coding, for instance, Zero-shot prompting might lead to suboptimal performance due to the absence of domain knowledge and contextual understanding.
In contrast, our CoI can integrate domain knowledge, enabling LLMs to pay attention to underlying information that are unavailable in the utterances.

\subsubsection{Few-Shot Prompting}
LLMs have been proven to have the capability of in-context learning, which makes Few-Shot Prompting~\cite{brown2020language,liu2023large} more effective than Zero-Shot Prompting. 
Before performing a specific task, Few-Shot prompting provides LLMs with a demonstration of examples (i.e., a small number of ``shots'') for tuning, including task examples and their corresponding ground truth outputs.
These examples serve as a brief learning guide, providing LLMs with some context or insight about the nature of the task.
However, Few-Shot prompting does not actively guide the model in breaking down the task into multiple reasoning steps. 
Therefore, for tasks requiring multi-step reasoning, such as basic mathematical problems~\cite{cobbe2021training} and the MI behavioral coding in this study, Few-shot prompting still struggles to perform well.
In contrast, our CoI prompting reduces the complexity of the behavioral coding task by breaking it down into three sequential sub-tasks and prompt stages.

\subsubsection{Zero-Shot Chain-of-Thought Prompting}
Marking a departure from traditional Chain-of-Thought~\cite{wei2022chain}, which typically relies on hand-crafted, detailed Few-shot examples for each task, Zero-shot Chain-of-Thought (ZeroCoT)~\cite{kojima2022large} is a template-based prompting method for LLMs.
It focuses on sequential reasoning without the need for specific training examples. 
By employing a generalized, one-size-fits-all prompt such as ``Let's think step by step''.
ZeroCoT effectively guides LLMs to decompose various complex tasks into multiple reasoning steps. 
Its robustness has been proven across a wide range of applications, from intricate arithmetic and symbolic reasoning problems to other logical reasoning tasks.
However, MI behavioral coding is not merely a task of logical reasoning. 
It also requires the use of domain-specific knowledge and the characteristics of interactions between patients and therapists to assist in coding.
For this reason, our CoI method, through three well-designed prompt stages, breaks the MI behavior coding task down into three sequential sub-tasks based on professional guidelines and coding schema, rather than letting LLMs decompose the task on their own as ZeroCoT.
This approach aids LLMs in understanding the interactions between patients and therapists and simulates the reasoning process of human coders.

\subsection{Models}
We performed experiments on four LLMs, namely Llama2-13B-Chat~\cite{touvron2023llama}, Falcon-7B-Instruct~\cite{almazrouei2023falcon}, Mistral-7B-Instruct~\cite{jiang2023mistral}, and ChatGPT~\cite{openai2022chatgpt}, which are available by OpenAI API\footnote{\url{https://platform.openai.com/docs/guides/gpt/chat-completions-api}} or Hugging Face\footnote{\url{https://huggingface.co/}}. 
The reason for selecting these LLMs over their non-fine-tuned base versions (e.g. Llama2-13B-Base, Falcon-7B-Base) is their superior ability to follow dialogue-style instructions.

\subsubsection{Llama2-13B-Chat}
Llama2~\cite{touvron2023llama} is an auto-regressive Transformer model pre-trained on publicly available online data.
Additionally, it utilizes ghost attention~\cite{bai2022constitutional} and grouped query attention to enhance consistency in multi-turn dialogues and extend its context length, respectively.
In this study, we use the Llama2-Chat variant, which undergoes supervised fine-tuning followed by reinforcement learning from human feedback (RLHF)~\cite{christiano2017reinforcement, Stiennon2020Learning, ouyang2022instructGPT}, which includes rejection sampling and proximal policy optimization. 
While enhancing safety, its performance remains competitive with other open-source LLMs.


\subsubsection{Falcon-7B-Instruct}
\cite{Guilherme2023refinedweb} demonstrated the intrinsic link between the performance of LLMs and the quality of their training data.
Consequently, the Falcon-7B-Instruct~\cite{almazrouei2023falcon}, known for its superior performance, was trained on the meticulously curated RefinedWeb~\cite{Guilherme2023refinedweb} dataset.
To further enhance its efficiency and reduce computational overhead, Falcon employs multi-query and flash attention~\cite{dao2022flashattention} mechanisms, enables it to support sequences up to 2,048 tokens.

\subsubsection{Mistral-7B-Instruct}
Mistral-7B-Instruct~\cite{jiang2023mistral}, an instruction-tuned version of the Mistral model checkpoint, leverages the capabilities of Grouped-Query Attention and Sliding Window Attention to expedite inference processes and handle longer sequences. 
Compared to LLMs with an equivalent number of parameters, Mistral performs better in various domains, including Commonsense Reasoning, Reading Comprehension, and Mathematics.

\subsubsection{ChatGPT}
ChatGPT~\cite{openai2022chatgpt} represents an evolution in the Generative Pre-trained Transformer (GPT) series, with a focus on conversational capabilities.
It showcases superior language understanding and generation abilities, excelling in benchmarks like SuperGLUE~\cite{wang2019superglue} and HumanEval\cite{chen2021evaluating}, which require deep contextual insight and logical coherence.
It also integrates reinforcement learning from human feedback (RLHF)~\cite{christiano2017reinforcement, Stiennon2020Learning, ouyang2022instructGPT}, allowing it to generate more contextually appropriate and human-like responses in dialogue scenarios. 

\subsection{Sampling Strategy and Experiment Details }

To ensure stable and high-quality output, we adhere to the default sampling strategies specific to each LLM. 
For ChatGPT, we utilize GPT-3.5-Turbo via its \textit{ChatCompletion} API,
employing  Nucleus Sampling~\cite{Holtzman2020The} with top-p and temperature parameters set to 1.
This method known for enhancing output diversity by selecting the next word from a restricted set of highly probable options, thus facilitating more creative and varied responses.
For all other LLMs, we employ Greedy Sampling, a strategy that selects the most probable next word, ensuring deterministic and predictable text generation.
Considering the different context length limitations of various LLMs, we established a fair comparison by setting the Few-Shot N value to 1, which means providing one example before each sample.
Additionally, the Few-shot examples were randomly chosen from the support set to mitigate the risk of data leakage.

\subsection{Answer Extraction}
The open-ended nature of language generation makes it very challenging to evaluate the performance of LLMs on classification tasks. 
Two generalized evaluation models are to convert the classification task into multiple-choice questions (MCQ) and then extract the probability of the first token to select the label, and the other is to extract the corresponding answer from the complete answer based on a regular expression.
Previous study\cite{wang2024my} demonstrated the MCQ pattern may not consistently reflect the final response output due to varying response modes of the models.
Therefore, we use regular expressions to extract the first output that conforms to the format in the third stage (Valence Analysis).
For example, if a model response is: ``The patient's valence should be coded as neutral or positive,'' we extract the first matching token ``neutral'' as the model prediction. 
If the models do not follow the instructions, leading to no tokens in the output matching any labels, in this case, we randomly select a label.
Furthermore, the models we selected are aligned to ensure their outputs are consistent with human professional guidelines. 
But, this occasionally triggers the Content Safety Policy when generating content related to psychotherapy. 
Therefore, we opted to exclude those samples that trigger this policy.
These special treatments reduce the errors caused by the uncertainty of the generated model and improve the reliability of the research.

\section{Results}

Due to the data imbalance shown in Table~\ref{table: Data_statistics}, we chose Micro-averaged F1 and Macro-averaged F1 scores for our evaluation. 
Micro-averaged F1 assesses overall accuracy, while Macro-averaged F1 handles label imbalance by giving equal weight to all classes.
These metrics offer a comprehensive view of our Chain-of-Interaction approach's performance, both for comparing it against baseline methods and for conducting an ablation study.

\begin{table*}[ht]
\centering
\caption{Main results for the behavioral coding task, expressed in percentages. We \textbf{bolden} the best performance.}
\label{table: Main results}
\begin{tabular}{l|cc|cc|cc|cc|cc}

\multirow{2}{*}{Methods (\%)} & \multicolumn{2}{c|}{Llama2}   & \multicolumn{2}{c|}{Falcon}   & \multicolumn{2}{c|}{Mistral}  & \multicolumn{2}{c|}{ChatGPT}  & \multicolumn{2}{c}{Average}  \\
                         & Micro-F1          & Macro-F1          & Micro-F1          & Macro-F1          & Micro-F1          & Macro-F1          & Micro-F1          & Macro-F1          & Micro-F1          & Macro-F1          \\ \hline \hline
Zeroshot                 & 42.7          & 33.6          & 40.3          & 31.0          & 39.6          & 30.4          & 52.0          & 38.0          & 43.7          & 33.3          \\
Fewshot                  & 46.5          & 33.8    & 43.9   & 31.1   & 47.3   & 33.6   & 51.9          & 38.8          & 47.4          & 34.3    \\
ZeroCoT                  & 62.6   & 33.6          & 40.4          & 30.9          & 44.9          & 31.5          & 55.6    & \textbf{41.0} & 50.9    & 34.3    \\
CoI                    & \textbf{63.5} & \textbf{34.5} & \textbf{53.2} & \textbf{32.6} & \textbf{52.0} & \textbf{34.6} & \textbf{60.2} & 40.5    & \textbf{57.2} & \textbf{35.6} \\ 
\end{tabular}
\end{table*}

\subsection{Main Results}
Table~\ref{table: Main results} presents performance results for different prompting methods applied to the MI behavioral coding task across various large language models (LLMs).
The results show that our CoI prompting significantly outperforms the state-of-the-art prompting baselines.
For example, our method outperforms the second-best approach, Zero-shot Chain-of-Thought\cite{kojima2022large}, in average performance across four LLMs. 
Specifically, it shows improvements of 6.3\% in Micro-F1 and 1.3\% in Macro-F1 score.
This superior performance can be attributed to our method's strategic breakdown of the behavioral coding task into three key stages that leverage domain knowledge. 
By segmenting the task into multiple interactive stages, our approach facilitates step-by-step reasoning and enables the model to navigate the complexities of the task in a structured and informed way. 
This progressive reasoning simulates the layered understanding that a psychological professional may employ, leading to more accurate predictions and a more profound interpretation of patient behaviors.

While different baseline prompting methods exhibit significant performance variations across various LLMs, our method consistently achieves state-of-the-art results. 
For instance, the performance of ZeroCoT may be comparable to our proposed CoI method on LLaMA2 but appears closer to the simpler Zero-shot prompting on Falcon, highlighting the performance instability of baseline methods.
Few-shot prompting improves performance over Zero-shot on three LLMs, but its impact on ChatGPT may be minimal, possibly suggesting its training may already include similar tasks. 
Despite this instability in baseline methods, our approach stands out by consistently outperforming them, demonstrating robustness and adaptability across different LLMs.

\subsection{Ablation Study}

\begin{table*}[ht]
\centering
\caption{Results of the Ablation Analysis: Utilizing Micro-F1, Macro-F1, expressed in percentages}
\begin{tabular}{l|cc|cc|cc|cc|cc}
\multirow{2}{*}{Methods (\%)} & \multicolumn{2}{c|}{Llama2}   & \multicolumn{2}{c|}{Falcon}   & \multicolumn{2}{c|}{Mistral}  & \multicolumn{2}{c|}{ChatGPT} & \multicolumn{2}{c}{Average}  \\
                         & Micro-F1          & Macro-F1          & Micro-F1          & Macro-F1          & Micro-F1          & Macro-F1          & Micro-F1           & Macro-F1        & Micro-F1          & Macro-F1          \\ \hline\hline
Zeroshot                 & 42.7          & 33.6          & 40.3          & 31.0          & 39.6          & 30.4          & 52.0           & 38.0        & 43.7          & 33.3          \\
w/o ID                   & 48.4          & 33.5          & 43.1          & 30.6          & 50.0          & 33.9          & 52.5           & 40.8        & 48.5          & 34.7          \\
w/o IA                   & 54.6          & 33.5          & 45.7          & 31.1          & 43.4          & 31.2          & 56.2           & 39.9        & 50.0          & 33.9          \\
w/o VA                   & 62.9          & 33.9          & 51.4          & 32.7          & 50.1          & 33.8          & 58.5           & 40.6        & 55.7          & 35.3          \\
CoI                     & 63.5 & 34.5 & 53.2 & 32.6 & 52.0 & 34.6 & 60.2 & 40.5  & 57.2 & 35.6 \\
\end{tabular}

\label{table: ablation-study-results}
\end{table*}


To systematically explore the contribution of each stage in our Chain-of-Interaction (CoI) method, we conduct a series of ablation studies and present the results in Table~\ref{table: ablation-study-results}.
These studies start with a baseline scenario, termed \textit{Zeroshot}, where all CoI stages are omitted. 
This baseline indicates the performance of models without the benefit of any structured CoI stages.
Subsequently, we gradually removed Interaction Definition stage (\textit{w/o ID}), Involvement Assessment stage (\textit{w/o IA}), and Valence Analysis stage (\textit{w/o VA}).

We summarize our results in Table~\ref{table: ablation-study-results}. Removing the Interaction Definition stage (\textit{w/o ID}) leads to LLMs beginning their subjective rating process about the patient's emotional expression and self-exploration, without fully understanding the therapist-patient interaction context by the guidance of MISC behavior coding schema. 
Compared to the full CoI method, Micro-F1 and Macro-F1 decrease by  8.7\% and 0.9\%, respectively, which reflect the impact of the loss in initial insight into the dynamics of the interaction.
When we exclude the Involvement Assessment stage (\textit{w/o IA}), the LLMs no longer mimic the subjective process used by MI professionals to assess patient involvement.
In this setting, LLMs analyze the patient's general sentiment and valence directly after reasoning about the specific patterns of interaction between the patient and the therapist. 
Valence was analyzed without the help of the patient's self-exploration and emotional expression, which are two key indicators of it.
The absence results in a decrease of 7.2\% in Micro-F1 and 1.7\% in Macro-F1 scores compared to the full method.
Finally, skipping the general sentiment analysis in the Valence Analysis stage (\textit{w/o VA}) deprives the model of fully indications of the patient's emotional state. 
This absence can affect the model's ability to accurately code patient behaviors, as it misses out on initial emotional cues that could influence the overall understanding of the patient's attitude towards change.
In this scenario, the average performance of the LLMs experiences a 1.5\% drop in Micro-F1 and a 0.3\% drop in Macro-F1 compared to the full method.

In conclusion, the full CoI method yields the best results, showcasing the method's effectiveness in leveraging domain knowledge to enable the model to perform detailed reasoning. 
Each stage contributes to a composite understanding, which is reflected in the model performance comparisons, confirming that the integrated stages of CoI are crucial for a nuanced task like MI behavioral coding.

\section{Related Work}

\subsection{Automatic Behavioral Coding}

Automatic behavior coding is a critical task to examine the fidelity of MI sessions which, if conducted in real time, can have valuable training and clinical applications.
Due to domain-expertise and time-consuming natures of the task, the common strategy is to develop a machine learning classifier to predict and assess patient behaviors in MI sessions, such as substance disorder~\cite{xiao2016technology}, suicide~\cite{levkovich2023suicide},  and alcohol addiction~\cite{huang2018modeling}.
In recent years, neural models have dominated the automatic behavioral coding task and showed their supremacy performance than non-neural approaches, such as linguistic features~\cite{can2012case} and topic models~\cite{atkins2014scaling}.
Existing studies~\cite{singla2018using, huang2018modeling, gibson2017attention, gibson2022multi-task-behavior-coding} develop end-to-end classification pipelines by encoding patient utterances into neural feature vectors and utilizing the vectors for predictions by neural classifiers, such as recurrent neural network (RNN).
For example, \cite{  huang2018modeling, cao2019observing, singla2020towards} deployed Long-Short Term Memory (LSTM) or Gated Recurrent Unit (GRU) to enrich contextual representations and \cite{gibson2017attention, singla2018using} extended the RNN variants (LSTM and GRU) with attention mechanisms, which identify salient words and patterns in utterances.
More recent studies~\cite{tran2023multimodal, gibson2022multi-task-behavior-coding} have switched to pre-trained language models (e.g., BERT~\cite{devlin2019bert}) as the data feature extractor and neural network classifier, which requires additional fine-tuning steps by the annotated MI corpora.

While the behavioral coding task has achieved promising performance, lacking explicit incorporation of domain knowledge is a major issue of the existing methods, which can lead to unreliable performance comparing to human professionals and get worse when a large amount of training data is not available.
In this study, we fill the domain knowledge issue, propose a new prompt learning approach, and employ LLMs for the automatic coding of motivational interviewing, which can enable classification models behave as domain experts and learn the domain knowledge required for MISC coding as instructions.

\subsection{Large Language Models for Mental Health}





The field of mental health research is witnessing significant advancements with the integration of LLMs, as illustrated by a series of pioneering studies~\cite{demszky2023using,he2023towards}.
Their comprehensive analysis discusses the potential and limitations of LLMs for mental health, covers technologies ranging from pre-training to instruction tuning and prompt tuning.

Consistent with their exploratory research, both instruction tuned and frozen LLMs have been demonstrated to be productive in the Mental health domain.
When high-quality data is available, instruction tuning~\cite{lou2024comprehensive,lou2024zeroshot} LLMs tends to significantly improve their ability to perform on relevant tasks.
For example, ChatCounselor~\cite{liu2023Chatcounselor} and ChatDoctor~\cite{li2023chatdoctor}, as consulting chatbots, leverage instruction-tuned open-source LLMs to achieve performance levels comparable to GPT-4~\cite{openai2023gpt4}, while requiring significantly fewer computational resources.
And~\cite{lai2023psy-llm} conducted instruction tuning on LLMs using real-world psychological Q\&A sessions, enabling these models to acquire psychological knowledge and enhance their capability to provide counseling services.
In scenarios where high-quality data is lacking, existing studies usually utilize state-of-the-art LLMs to generate synthetic data for fine-tuning their LLMs.
For instance~\cite{yang2023mentalllama} used ChatGPT to generate instructions for training open-source LLMs, the instruction-tuned models achieved state-of-the-art performance on mental health detection tasks.
Although the instruction tuned LLMs are intuitively more effective than the frozen LLMs, the frozen LLMs with special prompting methods~\cite{zhou2023thread,wei2022chain,qi2023supervised,chen2023empowering} can also perform well with mental health tasks, such as suicidal risk classification~\cite{qi2023supervised} and cognitive distortions detection~\cite{chen2023empowering}.
Specifically, \cite{qi2023supervised} employed three strategies to assess the performance of LLMs on suicidal risk classification, including Zero-shot prompting, Few-shot prompting and instruction tuning.
The results show that both prompting and instruction tuning can improve the reasoning ability of LLMs.
Moreover, \cite{chen2023empowering} introduced the Diagnosis of Thought (DoT) prompting framework, which is designed to strategically prompt LLMs to generate diagnosis rationales, with a particular focus on the detection of cognitive distortions.

Likewise, our Chain-of-Interaction method focuses on prompting frozen LLMs.
Our approach differs from the existing prompting methods of LLMs for mental health studies, including Zero Chain-of-Thought~\cite{kojima2022large} and \cite{brown2020language,liu2023large}.
Our approach deconstructs reasoning steps grounded in domain-specific knowledge and the interaction patterns between patient and therapist, while Zero Chain-of-Thought~\cite{kojima2022large} decomposes based on general knowledge and the Few-shot~\cite{brown2020language,liu2023large} method relies on in-context learning without considering domain knowledge derived from coding schema.
\section{Conclusion}

In this study, we focus on the task of automatic coding of patient utterances during MI sessions. 
LLMs have demonstrated excellent performance on many tasks, and the prompting method is a key factor in their performance.
Previous prompting methods have enabled LLMs to excel in tasks like elementary mathematics and other common sense reasoning, but they fall short in domain-specific tasks like coding patient behavior that require specialized knowledge
and emphasize the dyadic contexts.

To address these specific challenges, we propose the \textit{Chain-of-Interaction} prompting method, which aims to leverage domain knowledge and patient-therapist interactions to enhance reasoning in LLMs.
This method decomposes the task of coding patient utterances in MI into multiple key reasoning steps through three sequential stages. 
It leverages the interaction characteristics between patients and therapists, allowing the large language model to reason using psychological domain knowledge about behavioral coding without further training.

On a real-world dataset of motivational interviews, we compared our proposed Chain-of-Interaction prompting method with three other popular prompting methods using four advanced large language models. 
The results indicate that our method achieves state-of-the-art performance. 
We also conducted an ablation analysis, and the experimental results show that removing any stage of the Chain-of-Interaction leads to a significant decrease in performance, demonstrating the effectiveness of each stage.

\section{Limitation}
In this study, we focus on uni-modal large language models that are trained only on text, unlike multi-modal models like GPT-4V~\cite{openai2023gpt4} which can also process audio. 
This limitation means that our models cannot leverage audio features from therapy recordings that have not been transcribed. 
Additionally, the sensitive nature of privacy in psychotherapy presents significant challenges in obtaining experimental data. 
Therefore, our experiments were only conducted using data from a select group of college students who were enrolled in mandatory alcohol cessation interventions.

\section{Ethics Statement}
We access the raw data in accordance with data agreements and underwent required training. 
Out of respect for ethical and privacy concerns, we will not release any clinical data that could be linked to individual patients. 
Instead, we are dedicated to sharing our code and detailed guidelines to facilitate the replication of our results.
Our focus lies primarily on computational methodologies, and we do not engage in direct data collection from human subjects. Furthermore, our institution's review board has verified that our study does not require IRB approval.

\section{Acknowledgment}
We would like to thank Precious Jones (NSF REU under the project IIS-2245920, PI: Huang) for her contributions in preprocessing the data errors and extracting coding schema from MI coding manuals.
The codes analyzed were created in the NIH-NIAAA R01AA017427 project (PI: Borsari).
We also appreciate the insightful comments of the ICHI-24 anonymous reviewers.

\bibliographystyle{IEEEtran}  
\bibliography{DL4MI}     
\end{document}